\documentclass[a4paper]{article}

\usepackage{INTERSPEECH2018}

\usepackage{breqn}
\usepackage{graphicx}
\usepackage{subcaption}
\usepackage[font=footnotesize,labelfont=bf]{caption}
\usepackage[breaklinks=true,letterpaper=true,colorlinks,bookmarks=false]{hyperref}
\usepackage{booktabs}
\usepackage{diagbox} % Defines \backslashbox{..}{..}

\title{The Conversation: Deep Audio-Visual Speech Enhancement}
\name{Triantafyllos Afouras, Joon Son Chung, Andrew Zisserman
}
\address{
  Visual Geometry Group, Department of Engineering Science,\\
  University of Oxford, UK}
\email{\{afourast,joon,az\}@robots.ox.ac.uk}

\def\newpara{\vspace{2pt}}
\def\psubsec{\vspace{-5pt}}

\begin{document}
\maketitle
\begin{abstract}
Our goal is to isolate individual speakers from multi-talker simultaneous speech in  videos.
Existing works in this area have focussed on trying to separate
utterances from known  speakers in controlled
environments. In this paper, we propose a deep audio-visual speech
enhancement network that is able to separate a speaker's voice given
lip regions in the corresponding video, by predicting both the magnitude and the
phase of the target signal. The method is applicable to speakers unheard and unseen during
training, and for unconstrained environments. We demonstrate strong quantitative and
qualitative results,  isolating extremely challenging real-world
examples.
\end{abstract}


\section{Introduction}
\psubsec
In the film {\em The Conversation} (dir.\ Francis Ford Coppola, 1974),
the protagonist, played by Gene Hackman, goes to inordinate lengths to
record a couple's converservation in a crowded city square. Despite many
ingenious placements of microphones, he did not use the lip motion of
the speakers to suppress speech from others nearby.  In this paper we
propose a new model for this task of audio-visual speech enhancement, that he could
have used.

More generally,
 %the objective of this paper is to
%separate the speech signal of a target speaker from other speakers
%using visual information from the target speaker's lips. 
we propose an audio-visual neural network that can isolate a speaker's voice from others, using visual information from the target speaker's lips:
Given a noisy audio signal and the corresponding speaker video, 
we  produce an enhanced audio signal containing only the target speaker's voice with the
rest of the speakers and background noise suppressed.

%We consider scenarios where multiple speakers speak simultaneously.

\begin{figure}[!b]
      \centering
      \vspace{-17pt}
      \includegraphics[width=\columnwidth]{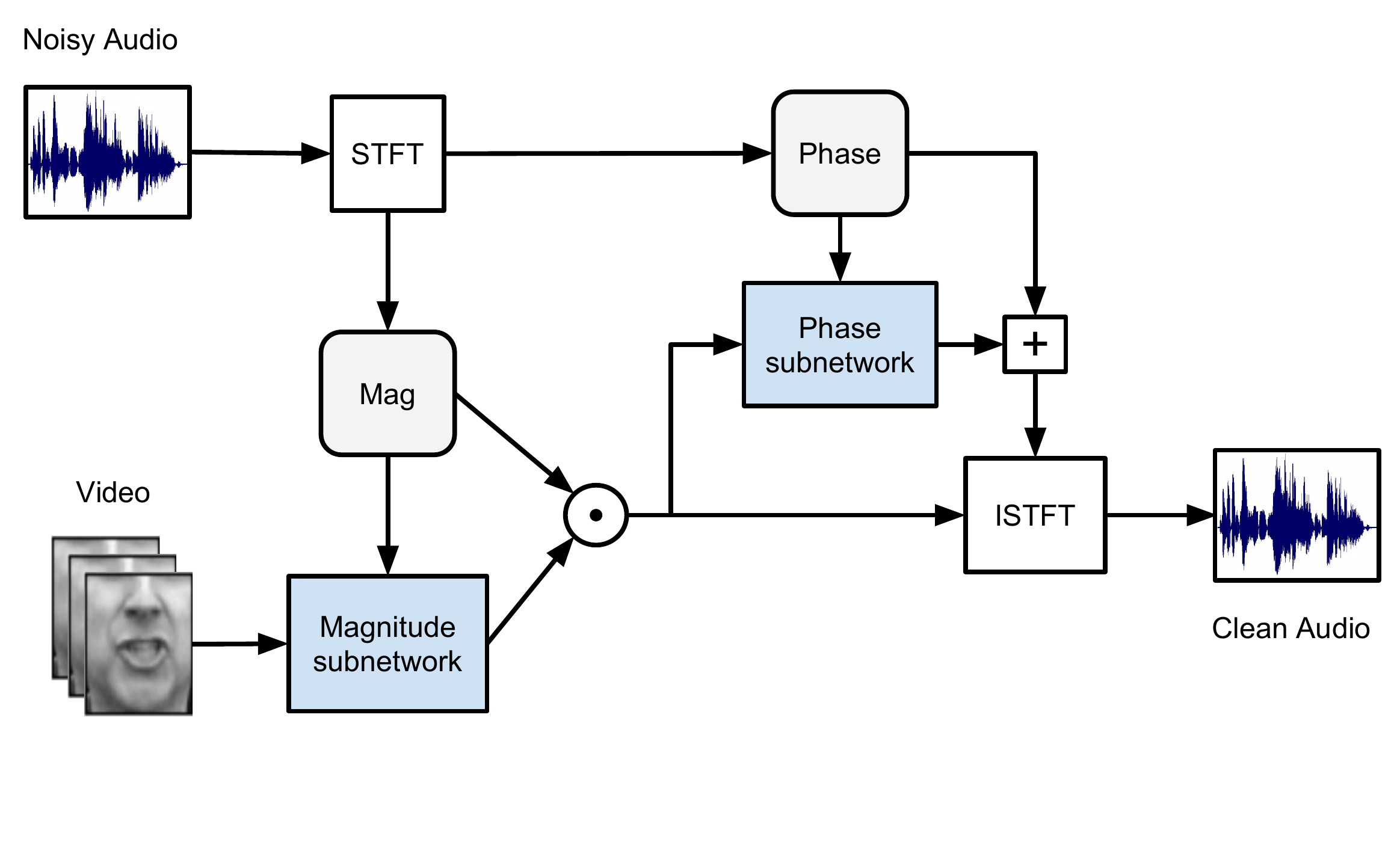}
      \vspace{-20pt}
      \caption{
       Audio-visual enhancement architecture overview.
It consists of two modules:  a  magnitude sub-network and a  phase sub-network.
The first sub-network 
receives the magnitude spectrograms of the noisy signal and the speaker video as inputs and
outputs a soft mask. We then multiply the input magnitudes element-wise with the mask to produce a
filtered magnitude spectrogram. 
The magnitude prediction, along with the phase spectrogram
obtained from the noisy signal are then fed into the second sub-network, which produces a phase
residual. The residual is added to the noisy phase, producing the enhanced phase spectrograms. 
Finally the enhanced magnitude and phase spectra are transformed back to the time domain, yielding
the enhanced signal.
}
      \label{fig:denoise_overview}
      \vspace{-15pt}
\end{figure}

Rather than synthesising the voice from scratch, which would be a challenging task, we instead predict a mask that
filters the noisy spectrogram of  the input.
Many speech enhancement approaches focus on refining only the magnitude of the noisy input signal and
use the noisy phase for the signal reconstruction. %~\cite{}.
This works well for high signal-to-noise-ratio scenarios, but as the SNR decreases, the noisy
phase becomes a bad approximation of the ground truth one~\cite{Fu2017}.  
Instead, we propose correction
 modules for both the magnitude and phase.
The architecture is summarised  in Figure~\ref{fig:denoise_overview}.
In training, we initialize the visual stream with a network pre-trained on a word-level lip-reading task, but
after this, we train from unlabelled data (Section~\ref{subsec:dataset}) where no explicit annotation 
is required  at the word, character or phoneme-level.

There are many possible applications of this model; one of them is automatic speeech recognition (ASR)
-- while machines can recognise speech relatively well in noiseless
environments, there is a significant deterioration in performance for
recognition in noisy environments~\cite{anusuya2010speech}. The
enhancement method we propose could address this problem, and improve,
for example, ASR for mobile phones in a crowded environment, or automatic
captioning for YouTube videos.

The performance of the model is evaluated for up to five  simultaneous
voices, and we demonstrate both strong qualitative and quantitative
performance. The trained model is evaluated on unconstrained 'in
the wild' environments, and for speakers and languages unseen at
training time.  To the best of our knowledge, we are the first to achieve
enhancement under such general conditions.
We provide supplementary material with interactive demonstrations on
\url{http://www.robots.ox.ac.uk/~vgg/demo/theconversation}.

%%  ---------- ---------- ---------- ---------- ---------- ---------- ---------- ----------
%%  ---------- ---------- ---------- ---------- ---------- ---------- ---------- ----------

\subsection{Related works}
\psubsec
Various works have proposed methods to isolate multi-talker simultaneous speech.
The majority of these are based on methods that only use the audio,
{\em e.g.} by using voice characteristics of a known
speaker~\cite{reddy2007soft,jin2009supervised,radfar2007single,makino2007blind,wang2017supervised}.
%In this paper, we instead focus on methods that make use of audio-visual information.
Compared to audio-only methods, we not only separate the
voices but also properly assign them to the speakers, by using the visual information.

Speech enhancement methods have traditionally only dealt with filtering the spectral magnitudes,
however many approaches have been recently been proposed for jointly enhancing the magnitude and  phase spectra~\cite{Fu2017, Mowlaee15, Mowlaee16,
Fahringer16, Williamson16, Hirsch17, Dubey17}.
%Joint magnitude and phase enhancement has been recently investigated by~\cite{Fu2017, Mowlaee15, Mowlaee16,
%Fahringer16, Williamson16, Hirsch17, Dubey17}.
The prevalent method for estimating phase spectra from given magnitudes in speech synthesis is
the one proposed by Griffin and Lim \cite{Griffin84}.

Prior to deep learning, a large number of previous works have been
developed for audio-visual speech enhancement by predicting
masks~\cite{liu2013source,khan2013speaker} or
otherwise~\cite{wang2005video,girin2001audio,deligne2002audio,hershey2002audio,hershey2004audio,Almajai09,Goecke02}, with 
an overview of audio-visual source separation is provided in~\cite{Rivet14}. However, we will concentrate from hereon on methods
that have built on these using a deep learning framework.

%The work of uses a deep neural network model 
%trained to synthesize audio from silent video~\cite{ephrat2017improved}. 
In ~\cite{ephrat2017improved} a deep neural network is developed to generate speech from silent video frames of a speaking person. 
This model is used in \cite{Gabbay17a} for speech enhancement, where the predicted spectrogram serves as
a mask to filter the noisy speech. 
However, the noisy audio signal is not used in the pipeline,
and the network is not trained for the task of speech enhancement.
In contrast, \cite{Gabbay17b} synthesizes  the clean signal
conditioning on both the mixed speech input and the input video. 
% The input is passed through an hourglass-like network that processes sets of 5 input frames at a
% time, along with their corresponding spectrograms.
\cite{Hou2017} also use a similar audio-visual fusion method, trained to both generate the clean signal and to reconstruct the video.
Both papers use the phase of the noisy input signal as an approximation for the 
clean phase. 
%Although this works for certain scenarios ({\em e.g.} only one extra speaker),
%the performance greatly deteriorates as the SNR decreases. 
%Furthermore, these methods are limited in
%that they are only demonstrated for a 
However, these methods are limited in
that they are only demonstrated under constrained
conditions ({\em e.g.} the utterances consist  of a fixed set of phrases in~\cite{Hou2017} ),  or for a 
small number of speakers  that have  been seen during training.
%Furthermore, these methods are limited in
%that they are only demonstrated under extremely constrained
%conditions ({\em e.g.} the utterances consisting of fixed set of phrases), or for a very
%small number of speakers that have  been seen during training.

Our method differs from these works in several ways:
%(i) It is fully convolutional and therefore applied continuously in time in a sliding-window manner
%on both the video frames and the spectrograms;
(i) we do not treat the spectrograms as images but as temporal signals with the frequency bins as
channels; this allows us to build a deeper network with a large number of parameters that trains fast;
(ii) we generate a soft mask for filtering instead of directly predicting the clean magnitudes, which we found to
be more effective; (iii) we include a phase enhancing sub-network; and, finally,
(iv) we demonstrate  on previously unheard (and unseen) speakers and on in-the-wild videos.

In concurrent and independent work, \cite{Ephrat18} develop a similar system, based on dilated
convolutions and a bidirectional LSTM, demonstrating good results in unconstrained environments, while \cite{Owens18} train a network for
audio-visual synchronisation and successfully use its features for speech separation.

The enhancement method proposed here is complementary 
to
%  other examples of 
% audio-visual learning, such as visual speech synthesis~\cite{Suwajanakorn17,Chung17b} and 
lip reading~\cite{Chung16,Assael16,Stafylakis17}, which has
also been shown to improve ASR performance in noisy environments~\cite{Chung17, Petridis18}.

%%  ---------- ---------- ---------- ---------- ---------- ---------- ---------- ----------
%%  ---------- ---------- ---------- ---------- ---------- ---------- ---------- ----------

\section{Architecture}
\begin{figure*}[!t]
	\centering
	\vspace{-30pt}
		\includegraphics[width=1.8\columnwidth]{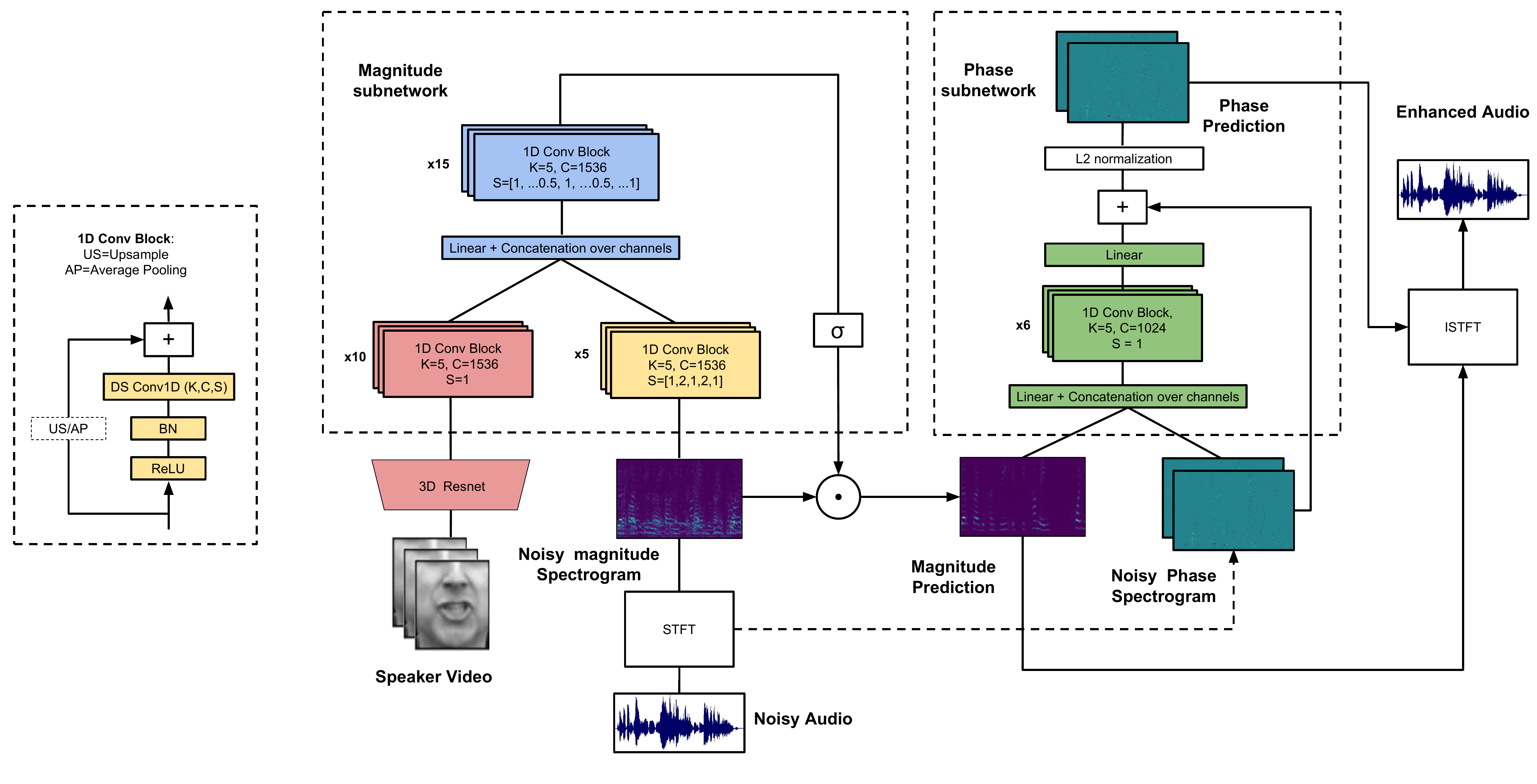}
		\vspace{-15pt}
                \caption{
                 Audio-visual enhancement network.  \textbf{BN}: Batch
                Normalization, {\bf C}: number of channels; {\bf K}: kernel width; {\bf S}: strides -- fractional ones denote transposed convolutions.
The network consists of a magnitude and a phase sub-network. 
The basic building unit is the temporal convolutional block with pre-activation~\cite{He16b} shown
on the left. Identity skip connections are added 
after every convolution layer (and speed up training).
All convolutional layers have $1536$ channels in the magnitude sub-network  and $1024$ in the phase 
subnetwork. Depth-wise separable convolution layers~\cite{chollet2017xception} are used, 
which consist of a separate convolution
along the time dimension for every channel, followed by a position-wise projection onto the new channel dimensions
(equivalent to a convolution with kernel width 1).
}
	 \label{fig:denoise_pipeline}
	 \vspace{-15pt}
\end{figure*}

\psubsec
This section describes the input representations and 
architectures for the audio-visual speech enhancement network.
The network ingests continuous clips of the audio-visual data.
The  model architecture is given in detail in Figure~\ref{fig:denoise_pipeline}.

\subsection {Video representation}
\psubsec
 
Visual features are extracted from the input image frame sequence with a spatio-temporal residual network similar to the one proposed 
by~\cite{Stafylakis17}, pre-trained on a word-level lip reading task.
The network consists of a 3D convolution layer, followed by a 18-layer ResNet~\cite{He15}.
%The resulting network is shown in Figure \ref{fig:frontend}.
%The kernel widths of the first convolutional layer are $5 \times 7 \times 7$ and are
%applied with a stride of $1$ on the temporal axis, with max pooling performed on the spatial
%dimensions only, so that the frame count of the input be preserved. 
For every video frame the network outputs a compact $512$ dimensional feature vector $f^v_0$
(where the subscript $0$ refers to the layer number
in the audio-visual network).
Since we train and evaluate on datasets with pre-cropped faces, we do not perform any extra pre-processing,
besides conversion to grayscale and an appropriate scaling.

\subsection {Audio representation}
\psubsec
The acoustic representation is extracted from the raw audio waveforms using  Short Time Fourier
Transform (STFT) with a Hann window function, 
which generates magnitude and phase spectrograms.
STFT parameters are computed in a similar manner to~\cite{Gabbay17b}, so that every
video frame of the input sequence corresponds to four temporal slices of the resulting spectrogram.
Since the videos are at 25fps (40ms per frame), we select a hop length of 10ms with a window
length of 40ms at a sample rate of 16Khz. The resulting spectrograms have frequency
resolution $F=321$, representing frequencies from 0 to 8 kHz, and time resolution $T\approx
\frac{T_{s}}{hop}$, where $T_s$ is the duration of the signal in seconds.
The magnitude and phase spectrograms are represented as
$T\times321$ and $T\times642$ tensors respectively, with the real and imaginary
components  concatenated along the frequency axis for the latter. We convert the magnitudes to
mel-scale spectrograms, with 80 frequency bins before feeding them to the magnitude subnetwork,
however we conduct the filtering on the original, linear-scale spectrograms.  

%Note: With audio an video networks having different depths, the video has larger receptive field.
%However we tried adding more layers to audio and it didn't have any significant speedup.

\subsection {Magnitude sub-network}
\psubsec
The visual feature sequence $f^v_{0}$ is processed by a residual network of
10 convolutional blocks.
Every block consists of a temporal convolution with kernel width 5 and stride 1, preceded by ReLU
activation and batch normalization. A shortcut connection adds the block's input to the result of
the convolution.
%The activation of the $l_{th}$ layer of the visual stream is:
%\begin{equation}
  %f^v_{l} = \text{BN}( \text{ReLu} ( \text{SepConv}( W^{K=5}_{S=1}, V_{l-1} ) ) )  +  f^v_{l-1} 
%\end{equation}
%where  with $\text{SepConv}( W^{K=5}_{S=1}$ we denote a temporal depth-separable convolution with kernel
%width $K$ and stride $K$.
A  similar stack of 5 convolutional blocks is employed for processing the audio
stream. The convolutions are performed along the temporal dimension, with the
frequencies of the noisy input spectrogram $M_n$ viewed as the channels. Two of the intermediate
blocks perform convolutions with stride 2, overall down-sampling the temporal dimension by 4, in order to bring it down to the
video stream resolution. The skip connections of those layers are down-sampled by average pooling
with stride 2.
%\begin{equation}
  %f^a_{l} = \text{BN}( \text{ ReLu} ( \text{ SepConv}( W^{K=5}_{S_l}, f^a_{l-1} ) ) )  +  f^a_{l-1}
%\end{equation}
%where $S_l = 2, l\in [2,4], \text{otherwise } 1, \text{ for } l \in [1,5]$. 
The audio and visual streams are then concatenated over the channel dimension: $f^{av}_0 =  [ f^v_{10} ; f^a_{5} ]$.
The fused tensor is passed through another stack of 15 temporal convolution blocks. Since we want the output mask to
have the same temporal resolution as the input magnitude spectrogram, we include two transposed convolutions,
 each up-sampling the temporal dimension by a factor of 2, resulting in a 
factor of $4$ in total.
%\begin{equation}
  %f^{av}_{l} = \text{BN}( \text{ ReLu} ( \text{ SepConv}( W^{K=5}_{S_l}, f^{av}_{l-1} ) ) )  +  f^{av}_{l-1}
%\end{equation}
%where $S_l = 0.5, l\in [5,10], \text{otherwise } 1, \text{ for } l \in [1,15]$. 
The fusion output is projected through position-wise convolutions onto the original magnitude spectrogram
dimensions and passed through sigmoid activation in order to output a mask with values between 0 and 1. The resulting
tensor is multiplied with the noisy magnitude spectrogram element-wise to produce the enhanced magnitudes: 
\vspace{-1pt}
\begin{equation}
  \hat{M} = \sigma ( W_m^{T}f^{av}_{15} ) \odot M_{n} \nonumber
\end{equation}

\subsection {Phase sub-network}
\psubsec
Our intuition for the design of the phase enhancement sub-network is that there is structure in speech 
that induces a correlation between the magnitude and phase spectrograms.
As with the magnitudes, instead of trying to predict the clean phase from scratch,
we only predict a residual that refines the noisy phase.  
The phase sub-network is therefore conditioned on both the noisy phase and the magnitude predictions.
These  two inputs are fused together through linear projection and concatenation
and then processed by a stack of 6 temporal convolution blocks, with 1024
channels each. The phase residual is formed by projecting the result onto the dimensions
of the phase spectrogram and is added to the noisy phase. The clean phase prediction is finally obtained by
$L_2$-normalizing the result:
\begin{align}
%& \phi_{0} =  \nonumber \\
  %& \phi_{l} = \text{BN}( \text{ReLu} ( \text{SepConv}( W^{K=5}_{S=1}, \phi_{l-1} ) ) )  +  \phi_{l-1}, l \in [1,6] \nonumber \\
%& \phi_{6} = 6\times ConvBlock([  W_{m\phi}^{T} \hat{M} ; W_{n\phi}^{T} \Phi_{n}  ]) \nonumber \\
& \phi_{6} =
\underbrace{ConvBlock( \dots  ConvBlock( [  W_{m\phi}^{T} \hat{M} ; W_{n\phi}^{T} \Phi_{n}  ]) )}_{\times6} \nonumber \\
  &\hat{\Phi} =  \frac{(W_\phi^{T} \phi_{6} + \Phi_{n} \nonumber) } { ||(W_\phi^{T} \phi_{6} +
  \Phi_{n})||_2} \nonumber 
\end{align}
In training, the weights of the layers are initialized with small values and zero biases,
so that the initial residuals are nearly zero and the noisy phase is propagated to the output.

\subsection{Loss function}
\psubsec
The magnitude subnetwork is trained by minimizing  the $L_1$ loss between the 
predicted magnitude spectrogram
and the ground truth. The phase subnetwork is trained by  maximizing the cosine similarity between the phase
prediction and ground truth, scaled by the ground truth magnitudes.
%Our intuition behind this choice is that the phase prediction error only needs  to be low at the
%high-energy frequency areas.
%** is this right? can you explain more where it matters if the 
%phase prediction error is low, and where it doesn't matter **
The overall optimisation objective is:
 \begin{equation}
   \mathcal{L} = ||\hat{M} - M^*||_{1} \ - \ \lambda \ \frac{1}{TF}  \sum_{t,f} M_{tf}^* 
<\hat{\Phi}_{tf}, \Phi^*_{tf}>
 \end{equation}

%%  ---------- ---------- ---------- ---------- ---------- ---------- ---------- ----------
%%  ---------- ---------- ---------- ---------- ---------- ---------- ---------- ----------

\section{Experiments}
\psubsec
\begin{table*}[!t] 
\vspace{-20pt}
\begin{center}
\scriptsize
\begin{tabular}{ ll @{\hspace{3\tabcolsep}} rrrr @{\hspace{3\tabcolsep}} rrrr @{\hspace{3\tabcolsep}} rrrr @{\hspace{3\tabcolsep}} rrrr } 
 \toprule

 &  &  \multicolumn{4}{c}{SIR (dB)} & \multicolumn{4}{c}{SDR (dB)} & \multicolumn{4}{c}{PESQ} &\multicolumn{4}{c}{WER (\%)} \\  
\addlinespace[2pt]
%\multicolumn{2}{c}{\# Spk. }

Mag & \backslashbox[13mm]{$\Phi$}{\# Spk.}  & 2 & 3 & 4 & 5  & 2 & 3 & 4 & 5   &2 & 3 & 4 & 5  &2 & 3 & 4 & 5  \\ 
\cmidrule(l{-0.5\tabcolsep} r{2.5\tabcolsep}){3-6}  
\cmidrule(l{-0.5\tabcolsep} r{2.5\tabcolsep}){7-10}  
\cmidrule(l{-0.5\tabcolsep} r{2.5\tabcolsep}){11-14}  
\cmidrule(l{-0.5\tabcolsep} r{0\tabcolsep}){15-18}  
\cmidrule(r{2\tabcolsep}){1-2}  
{\bf Mix} & {\bf Mix}  &   $-$ & $-$  & $-$  &$-$      & -0.3 & -3.4 & -5.4  &-6.7     & 1.73  & 1.47  & 1.37 & 1.21 &     93.1 & 99.5 & 99.9 &100 \\
{\bf Pr} & {\bf GT}    &  10.8 & 13.2 & 13.8 &13.7     & 15.7 & 13.0 &  10.8 & 9.5     & 3.41  & 3.05  & 2.93 & 2.80 &     9.4  & 12.0 & 16.7 & 21.5 \\ 
 {\bf Pr} & {\bf GL}   &   0.9 & 2.5  & 3.6  &4.0      & -2.9 & -2.8 &  -2.9 &-2.7     & 2.98  & 2.71  & 2.52 & 2.35 &     10.5 & 13.7 & 20.3 & 27.8 \\
{\bf Pr} & {\bf Mix}   &   1.6 & 2.7  & 2.5  &2.0      & 10.5 &  7.8 &  5.9  & 4.8     & 3.02  & 2.70  & 2.49 & 2.33 &     10.8 & 14.9 & 22.0 & 31.9 \\ 
 {\bf Pr} & {\bf Pr}   &   3.9 & 5.4  & 5.4  &4.8      & 11.8 &  9.1 &  7.1  & 5.8     & 3.08  & 2.79  & 2.56 & 2.43 &     9.7  & 13.8 & 20.3 & 28.9 \\ 
%\cmidrule(l{1\tabcolsep} r{3\tabcolsep}){3-6}  
%\cmidrule(l{0\tabcolsep} r{3\tabcolsep}){7-10}  
%\cmidrule(l{0\tabcolsep} r{3\tabcolsep}){11-14}  
%\cmidrule(l{0\tabcolsep} r{\tabcolsep}){15-18}  
%\addlinespace[1pt]
%{\bf GT} & {\bf GT}  &   \multicolumn{4}{c}{Inf.} & \multicolumn{4}{c}{Inf.} & \multicolumn{4}{c}{Inf.} & \multicolumn{4}{c}{8.8} \\  
 \bottomrule

\end{tabular}    
\normalsize       
\vspace{-8pt}                                              
\end{center}
\caption{ 
  Evaluation of speech enhancement performance on the LRS2 dataset, for scenarios with different
  number of speakers (denoted by \# Spk).
The magnitude (Mag) and phase ($\Phi$) columns specify if the spectrograms used for the
reconstructions are predicted or are obtained directly from the mixed or ground truth signal:
{\bf Mix:} Mixed; {\bf Pr:} Predicted; {\bf GT:} Ground Truth; {\bf GL:} Griffin-Lim; 
{\bf SIR:} Signal to Interference Ratio; {\bf SDR:} Signal to Distortion Ratio; {\bf PESQ:}
Perceptual Evaluation of Speech Quality, varies between 0 and 4.5;
(higher is better for all three); {\bf WER:} Word Error Rate from off-the-shelf ASR system
(lower is better). The WER on the ground truth signal is $8.8\%$.  }
\label{tab:results}
\vspace{-25pt}
\end{table*}

\subsection{Datasets}
\label{subsec:dataset}
\psubsec
The model is trained on two datasets: the first is the BBC-Oxford
Lip Reading Sentences 2 (LRS2) dataset~\cite{Chung17,Chung17a}, which contains thousands of sentences
from BBC programs such as Doctors and EastEnders;
the second is VoxCeleb2~\cite{Chung18b}, which contains over a million utterances
spoken by over 6,000 different speakers. 
%The statistics of the 
%datasets is given in Table~\ref{}.

The LRS2 dataset is divided into training and test sets by broadcast date, in order to ensure that there is no overlapping video between the sets. The dataset covers a large number of speakers, which encourages the trained model to be speaker agnostic. However, since no identity labels are provided with the dataset, there may be some overlapping speakers between the sets. The ground truth transcriptions are provided with the dataset, which allows us to perform quantitative tests on the intelligibility of the generated audio.

The VoxCeleb2 dataset lacks the text transcriptions, however the dataset is divided into training and test sets by identity, which allows us to test the model explicitly  for speaker-independent performance.

The audio and video on these datasets are properly synchronized.
Evaluation on videos where this is not the case ({\em e.g.} TV broadcast), 
is possible by preprocessing with the pipeline described in \cite{Chung16a}
to detect and track active speakers and synchronize the video and the audio.

\vspace{-2mm}
\subsection{Experimental setup}
\psubsec

We examine scenarios where we add 1 to 4 extra interference speakers on the clean signal,
therefore we generate signals with 2 to 5 speakers in total.
It should be noted that the task of separating the voice of multiple speakers with equal average
``loudness'' is more challenging than separating the speech signal from background babble noise.

\vspace{-2mm}
\subsection{Evaluation protocol}
\psubsec
We evaluate the enhancement performance of the model in terms of
perceptual speech quality using the blind source separation criteria described in ~\cite{Fevotte05}
(we use the implementation provided by ~\cite{Emiya11}).
%The metrics are derived by decomposing the noise present in the enhanced signal into components accounting
%for different deformations and measuring energy ratios between them. 
The Signal to Interference Ratio (SIR) measures how well the unwanted signals have been suppressed,
the Signal to Artefacts Ratio (SAR) accounts for the introduction of artefacts by the enhancement
process,  and the Signal to Distortion Ratio (SDR) is an overall quality measure, taking both into account.
We also report results on PESQ \cite{Rix01}, which measures the overall perceptual quality and STOI \cite{Taal11},
 which is correlated with the intelligibility of the signal.
From the metrics presented above, PESQ has been shown to be the one correlating best with 
 listening tests that account for phase distortion\cite{Mowlaee15b}. 

Additionally, we use an ASR system to test for the intelligibility of
the enhanced speech. For this, we use the Google Speech Recognition
interface, and  report the Word Error Rates (WER) on the clean, mixed
and generated audio samples.

\vspace{-2mm}
\subsection{Training}
\psubsec
We pre-train the spatio-temporal visual front-end on a word-level lip reading task,
following~\cite{Stafylakis17}. This proceeds in two stages: first, training on 
the LRW dataset~\cite{Chung16}, which covers near-frontal poses;
and then on an internal multi-view dataset of a similar size. 
To accelerate the subsequent training process, we freeze the front-end, pre-compute and save the visual features for all the videos,
and also compute and save the magnitude and phase spectrograms
for both the clean and noise audio. 

Training takes place in three phases: first, the magnitude prediction sub-network is trained,
following a curriculum which starts with high SNR inputs (i.e.\  only one additional speaker) and
then progressively moves  to more challenging examples with a greater number of speakers;
second, the magnitude sub-network is frozen, and  only the phase network is trained ; finally,
the whole network is fine-tuned 
end-to-end. We did not experiment with the hyperparameter balancing the magnitude and phase loss terms, but set it to $\lambda=1$.

To generate training examples we first select a reference pair of visual and audio features ($v_r$, $a_r$) by
randomly sampling a 60-frame clean segment, making sure that the audio
and visual features correspond and are correctly aligned.
We then sample $N$ noise spectrograms $x_n, n\in[1,N]$, and mix them with the reference spectrogram in the frequency domain
by summing up the complex spectra, obtaining the mixed spectrogram $a_m$.
This is a natural way to augment our training data since a different combination of noisy audio
signals is sampled every time.
Before adding in the noise samples, we normalize their energy to have the
reference signal's one:   
%\begin{equation}
$a_m = a_r + \sum_n \frac{rms(x_r)}{rms(a_n)} a_n \nonumber$.
%\end{equation}

\subsection{Results}
\psubsec
\newpara\noindent\textbf{LRS2}.
We summarize our results on the test set of the LRS2 dataset in Table~\ref{tab:results}.
The performance under the different metrics is listed for the following signal types: 
The mixed signal which serves as a baseline, and the reconstructions that
are obtained using the magnitudes predicted by our network and either 
the ground truth phase, the phase approximated with the Griffin Lim algorithm, the mixed signal phase or the predicted phase.
The signal reconstructed from predicted magnitudes and phases is what we consider the final output of
our network.

The evaluation when using the ground truth phase is included as an upper bound to the phase prediction.
As can be seen from all measures on the mixed signal, the task becomes increasingly difficult as more speakers are  added. 
In general both the BSS metrics and PESQ correlate well with our observations.
It is interesting to note that while more speakers are added, the SIR stays
roughly the same, however more overall distortion is introduced.
The model is very effective in suppressing cross-talk in the output, however it does so with a trade-off in the quality of the target voice.

The phase predicted by our network performs better than the mixed phase.
Even though the improvement is relatively small in numbers, the difference in speech quality is noticeable as
the ``robotic'' effect of having off-sync harmonics is significantly reduced.  
We encourage the reader to listen to the samples in the supplementary material, where those
differences can be understood better. 
However, the considerable gap with the performance of the ground truth phase shows
that there is much room for improvement in the phase network.

The transcription results using the Google ASR are also in line with these findings. 
In particular, it is noteworthy that our model is able to generate highly intelligible results
from noisy audio that is incomprehensible by a human or an ASR system.

Although the content is mainly carried by the magnitude,
we see major improvement in terms of WER when using a better phase approximation.
It is interesting to note that, although the phase obtained using the Griffin Lim (GL) algorithm achieves
significantly worse performance on the objective measures,
it demonstrates relatively strong WER results, even slightly surpassing the predicted phase by a small margin in the case of 5 simultaneous speakers.

\newpara\noindent\textbf{VoxCeleb2.}
\vspace{-1pt}                              
%Achieving good enhancement quality on the LRS2 dataset showcases that our model can generalise across a large number of speakers.
In order to explicitly assess whether our model can generalize to speakers unseen during training,
we also fine-tune and test on VoxCeleb2, using train and test sets that are disjoint in terms
of speaker identities.
The results are summarized in Table~\ref{tab:results_vox}, where we showcase an experiment for the 3-speaker
scenario. We additionally include evaluation using the SAR and STOI metrics.
%one with a model that has only been trained on the LRS2 dataset, and one where it has been fine-tuned on VoxCeleb2.
Overall the performance is comparable to, but slightly worse than, on the LRS2 dataset --  which is in line with the qualitative performance.
This can be attributed to the visual features not being fine-tuned, and the presence of a lot of other background noise in VoxCeleb2.
The results confirm that the method can generalize to unseen (and unheard) speakers.

The last column of the table shows the PESQ evaluation for the original model trained on LRS2, without any fine-tuning on VoxCeleb.
The performance is worse than that of the fine-tuned model,  
however it clearly works. Since LRS2 is constrained to English speakers only,
but VoxCeleb2 contains multiple languages,
this demonstrates that the model learns to generalise to languages not seen during training. 

\begin{table}[h] 
\vspace{-6pt}                              
\begin{center}
\scriptsize
\begin{tabular}{ ll r r r r r| r r } 
 \toprule
          Mag  & $\Phi$ &  SIR   & SAR    & SDR   & STOI & PESQ  & PESQ-NF \\  
 \midrule
 {\bf Mix} & {\bf Mix}  &  -     & -1.59  & -2.99 & 0.34 & 1.58 & 1.58  \\  
 {\bf Pr} & {\bf GT}    &  11.43 & 16.41  & 10.30 & 0.77 & 3.02 & 2.79  \\ 
 {\bf Pr} & {\bf GL}    &  2.05  & 3.49   & -2.42 & 0.65 & 2.59 & 2.39  \\ 
{\bf Pr} & {\bf Mix}    &  1.72  & 13.54  &  6.71 & 0.65 & 2.59 & 2.41  \\ 
 {\bf Pr} & {\bf Pr}    &  5.02  & 13.77  &  7.91 & 0.67 & 2.67 & 2.45  \\ 
 \bottomrule

\end{tabular}    
\normalsize                 
\vspace{-8pt}                              
\end{center}
\caption{ 
Evaluation of speech enhancement performance on the VoxCeleb2 dataset, for 3 simultaneous speakers, 
 Notations are described in the caption of Table~\ref{tab:results}. Additional metrics used here: {\bf SAR:} Signal to Artefacts Ratio; {\bf STOI:}
Short-Time Objective Intelligibility, varies between 0 and 1; {\bf PESQ-NF:} PESQ score with a model that has not been fine-tuned on VoxCeleb;
Higher is better for all. }
\label{tab:results_vox}
\vspace{-25pt}
\end{table}

\subsection{Discussion}
\psubsec
\newpara\noindent{\textbf{Phase refinement.}}
%Estimating the phase from imperfect (predicted) magnitudes is much more challenging compared to estimating from real (ground truth) magnitudes. 
%Our phase enhancement subnetwork works very well when trained and evaluated on the ground truth
%magnitudes, but it is much harder to generalize when fed with the imperfect predictions;
%we noticed that this is also the case with using Griffin Lim. 
Training our whole network end-to-end decreases the phase loss and this might suggest that the inclusion of visual
features also improves the phase enhancement. However,  a thorough investigation to determine
if,  and to what extent,  this is true is left to future work. 
%in which case the results are indistinguishable by the human ear.
%Similarly,  Griffin Lim also works much better when applied to magnitudes of a real signal.
% It is clear that 

\newpara\noindent{\textbf{AV synchronization.}}
Our method is very sensitive to the temporal alignment between the
voice and the video.  We use SyncNet for the alignment, but since the
method can fail under extreme noise, we need to build some invariance
in the model.  
% Since the samples are artificially created, the signal
%  is clean when we sync it so that's not reflected.  
In future work this
will be incorporated in the model.

%\newpara\noindent{\textbf{Visual features.}}
%We are able to train models that perform very well on the speech enhancement task, without  
%The visual features obtained by training on word classification from lip reading, retain a lot of
%frame-wise information. To verify this we some experiments on voice-synthesizing, trying to
%reconstruct. 
%The expressiveness of the features is due 

%We considered treating the spectrograms as images and using 2D convolutions in order to exploit the
%local structure in frequency as well. 
%However as we need to preserve both the temporal and frequency dimensions, deep architectures with
%a large enough number of parameters for sufficient capacity have prohibitively large activations to fit in GPU memory.
%Exploring hourglass-like components such as the ones used for segmentation is left for future work. 

%%  ---------- ---------- ---------- ---------- ---------- ---------- ---------- ----------
%%  ---------- ---------- ---------- ---------- ---------- ---------- ---------- ----------

%\begin{figure}
  %\begin{subfigure}{0.3\columnwidth}
  %\centering
  %\includegraphics[]{figures/mix.pdf}
  %\caption{1a}
  %\label{fig:sfig1}
%\end{subfigure}%
%\begin{subfigure}{0.3\columnwidth}
  %\centering
  %\includegraphics[]{figures/pred.pdf}
  %\caption{1b}
  %\label{fig:sfig2}
%\end{subfigure}
%\caption{plots of....}
%\label{fig:spects}
%\end{figure}

%%  ---------- ---------- ---------- ---------- ---------- ---------- ---------- ----------
%%  ---------- ---------- ---------- ---------- ---------- ---------- ---------- ----------
\section{Conclusion}
\psubsec
In this paper, we have proposed a method to separate the speech signal of a target speaker from background noise and other speakers
using visual information from the target speaker's lips. The deep network produces realistic speech segments by predicting both the phase and the magnitude of the target signal; we have also  demonstrated that the network is able to generate intelligible speech from very noisy audio segments recorded in unconstrained `in the wild' environments.

%  add to final version + CDT acknowledgement
\vspace{5pt}
\newpara\noindent\textbf{Acknowledgements.}
Funding for this research is provided by the UK EPSRC
CDT in Autonomous Intelligent Machines and Systems, 
the Oxford-Google DeepMind Graduate Scholarship, and the EPSRC 
Programme Grant Seebibyte EP/M013774/1. We would like to thank Ankush Gupta
for helpful comments.

\clearpage

\bibliographystyle{IEEEtran}

\bibliography{shortstrings,vgg_other,mybib,vgg_local}

\end{document}